# Benchmarking Large Language Models on *Heterogeneous Cognitive and Biomedical Knowledge Graphs: The Role of Graph Grounding in Multi-Hop Reasoning and Hallucination Mitigation*

Ali Sarabadani [a] [0000-0002-3521-1401], Kheirolah Rahsepar Fard [a] [0000-0001-5452-4596] and Tahereh Jeirani [b]

[a] Department of Computer Engineering and Information Technology, University of Qom, Qom, Iran.

[b] Master of Science in Positive psychology, Qom, Iran

a.sarabadani@stu.qom.ac.ir, Corresponding author: rahsepar@qom.ac.ir, jeirani1386@gmail.com

ABSTRACT

*The advancement of large language models (LLMs) in biomedical and cognitive neuroscience domains is rapidly happened. However, their capacity to reliably reason over heterogeneous and multi-relational knowledge graphs (KGs) remains to be understand.*

*This research presents a benchmark of five state of the art LLMs-GPT 5.1, Gemini 2.0 Pro, Llama 3, Qwen, and BioGPT-across five diverse KGs: Drug Bank, NeuroSynth, Cognitive Atlas, Brain Map, and GDPKG LLM (Integrating Gene, Disease, and Pharmacogenomics Knowledge Graphs for Cognitive Neuroscience Using Large Language Models). A new multitier query dataset is introduced, designed to test both how deeply these models can reason and how factually accurate they are under two different conditions: KG- aware (with graph grounding) and KG- agnostic.*

*Inter annotator agreement was assessed using pairwise Cohen's Kappa and Fleiss' Kappa. For Cohen's Kappa the coefficient $\kappa = 0.781$, range 0.753-0.807 and for Fleiss' Kappa the coefficient is $\kappa = 0.77$. These values indicate substantial to almost perfect agreement and confirming the robustness of our annotation framework. Our results conclude that the following three conclusions: First, general purpose frontier models (GPT 5.1 and Gemini 2.0 Pro) consistently outperform domain specific models (BioGPT and Llama 3) on multi hop and high-level reasoning tasks, yet they exhibit a pronounced performance degradation from single hop to multi hop queries, underscoring the inherent difficulty LLMs face when traversing relational chains across cross domain entities. Second, KG-aware execution reduced hallucination rates across all models. Also, general- purpose models were less likely to hallucinate than do-main- specific ones when they had access to structured knowledge. Third, contrary to our initial hypothesis, domain specific models did not surpass general purpose models even in specialized tasks*

*without KG grounding, suggesting that the breadth of pretraining data in frontier models confers an unexpected advantage.*

*In this research practical guidelines for selecting LLM-KG integration strategies based on task complexity and computational constraints are presented. Hybrid neuro- symbolic architectures that combine domain- general reasoning with domain- specific knowledge are also recommended. Our benchmark lays the groundwork for future research and supports applications in personalized medicine, cognitive disorder diagnosis, and explainable AI.*

# 1. Introduction

Large language models (LLMs) have changed the way we access and reason about scientific knowledge. Transformer based architectures (Vaswani et al., 2017) and large scale pretrained models such as GPT-style systems (Brown et al., 2020; Raffel et al., 2020) have demonstrated remarkable capabilities in natural language understanding, few shot learning, and compositional reasoning. Researchers are moving beyond open- domain QA to see how LLMs handle structured knowledge, like knowledge graphs. Knowledge graphs are essential in neuroscience and biomedical branch. Large- scale biomedical KGs are used for drug–target analysis (Wishart et al., 2018), disease–gene association modeling (Himmelstein et al., 2017), and multi- modal data integration (Zhang et al., 2025). In cognitive neuroscience, resources such as NeuroSynth (Yarkoni et al., 2011), Brain Map (Laird et al., 2005), and Cognitive Atlas (Poldrack et al., 2011) represent distinct but complementary knowledge modalities, ranging from activation co-ordinates to structured cognitive ontologies. However, these graphs differ in schema design, entity granularity, and relation semantics, making cross-resource reasoning non trivial. The integration of LLMs with KGs has emerged as a promising paradigm for neuro-symbolic systems (Garcez et al., 2015). Recent research has focused on using LLMs to translate natural language directly into structured queries like SPARQL or Cypher (D'Abramo et al., 2025; Ma et al., 2025). Other approaches combine retrieval-augmented generation with graph-based grounding mechanisms to reduce factual hallucination (Lewis et al., 2020; Baek et al., 2020). These methods work well in controlled settings, but their performance across different domain- specific knowledge graphs has not yet been explored. Most prior work focuses on single graph benchmarks or general-domain datasets rather than scientifically complex domains such as cognitive neuroscience. Cognitive neuroscience is a challenging domain for evaluating how LLMs interact with knowledge graphs. Queries often require multi-hop reasoning across hierarchical cognitive con-structs, brain regions, receptor pathways, and pharmacological mechanisms. For instance, tracking a cognitive process to its neural bases and then to drug effects needs inference across different types of relations. Although multi-hop reasoning has been studied in open-domain datasets such as HotpotQA (Yang et al., 2018) and Complex Web Questions (Talmor and Berant, 2018). These benchmarks do not match validation KGs. Consequently, models may produce plausible but unsupported reasoning paths (Creswell and Shanahan, 2022; Ji et al., 2023). Another central limitation of LLMs is hallucination, de-fined as generating factually unsupported or fabricated information (Ji et al., 2023). In biomedicine and neuroscience, when models hallucinate, they can misrepresent what's already known or even make up non- existent associations. Grounding strategies reduce hallucination risk. However few studies compare grounded (KG-aware) and ungrounded (KG-agnostic) regimes under an integrated evaluation. Such comparative analysis is essential to understand whether performance gains stem from internal parametric knowledge or from structured graph constraints. To address these gaps, we introduce a multi-KG benchmarking frame-work for evaluating LLMs on cognitive neuroscience and related biomedical knowledge graphs. The benchmark spans five heterogeneous resources: Drug Bank (Wishart et al., 2018), NeuroSynth (Yarkoni et al., 2011), Cognitive Atlas (Poldrack et al., 2011), Brain Map (Laird et al., 2005), and GDPKG-LLM (RahseparFard et al., 2025). We construct a stratified query set consisting of single-hop factual queries, multi-hop relational queries, and higher-level reasoning tasks requiring integration of multiple graph com-ponents. Crucially, we evaluate models under two complementary regimes: a KG-aware mode in which LLMs translate natural-language questions into executable structured queries, and a KG-agnostic mode in which models answer directly from their internal representations. This dual- mode setup allows us assess translation, semantic matching, reasoning, and hallucination. By systematically comparing multiple frontier and domain-specific LLMs across heterogeneous KGs, we provide insights into the suit-ability of current models for neuro-symbolic cognitive neuroscience applications. The hypothesis is that frontier general purpose models would excel in surface level factual retrieval, domain specific models would demonstrate superior stability in complex reasoning tasks when provided with explicit graph grounding. Three experts independently coded hallucination and reasoning failures based on a qualitative error scheme. These results help build more LLM–KG pipelines for discovering knowledge, generating hypotheses, and reasoning structured scientific data.

## 2. Related Work

Research on large language models (LLMs) and structured knowledge graphs (KGs) spans several partially overlapping domains. It is included neural semantic parsing, graph-augmented generation, biomedical knowledge integration, and question answering benchmarks.

We propose a framework of knowledge-enabled LLM to solve the issue of few-shot retrieval, which adopts knowledge base to fully utilize the potential of LLMs in retrieval. We integrate KG-neuro-symbolic into the LLM approach via two knowledge enhancement modalities: i) leveraging KG's knowledge rendering ability, we generate a knowledge-enhanced prompt (KEP) to guide LLMs match queries and passages; ii) through KG's knowledge base, we expand the support set and patch the missing KG facts by performing cross-modal chain of reasoning (CoR). By incorporating knowledge and reasoning skills, our framework can improve the situation in few-shot passage and document retrieval, query passage similarity and text entailment, and text summarization with questions as passages. The extensive experimental results indicate that the proposed LLM-based solution with multi-modal knowledge augmentation is an efficient and effective strategy.

## 2.1. LLM-Based Structured Query Generation

As in prior studies, the LLM-based KG semantic parsing for natural language question answering (D'Abramo et al., 2025; Yin et al., 2016), and its application in deep learning systems (Srivastava et al., 2021) are summarized in this section. In traditional approaches, neural SPARQL parsing models that translate natural language questions into executable logical forms (Yin et al., 2016) were first introduced for KG semantic parsing. The deep neural models are further enhanced in the reinforcement learning-based neural SEQ-to-SEQ model and its hybrid COARSE-to-FINE architecture (Srivastava et al., 2021) to enable multi-hop inference and traverse graph paths. The early works mainly rely on multiple components in the pipeline to train neural models. However, the neural models highly rely on the selection of the KG schema and are usually unstable under different environments. Recent methods adopt the structure and features of KG to improve neural models for semantic parsing, e.g., the type/role-encoded word embedding, graph features, graph neural networks, graph-representation encoder, and multi-modality attention (Sarabadani et al., 2026). Moreover, a graph attention network is further introduced to understand the QA semantics ... The above methods heavily rely on the KG schema, resulting in instability and difficulty in performance optimization. The large number of parameters in the neural architecture and the computational complexity of learning are also challenges for these methods. Different incorporating external knowledge sources enables LLMs to reduce hallucination and improve factual grounding. Retrieval Augmented Generation (RAG) (Lewis et al., 2020) first retrieves relevant knowledge from external sources and subsequently generates an answer with a fitted LLM. However, all works only retrieve documents, which do not take full advantage of structured information such as gene-disease association. Recent works bring structures into the generation, such as retrieving triples, which requires the LLM to traverse a knowledge graph (Baek et al., 2020). The research on prompt-guided LLM based knowledge graph interaction is very promising, but to our knowledge, all works only retrieve general-domain knowledge or structures. Cross-KG reasoning and maintaining consistency across multiple knee structures are non-trivial.

## 2.2. Biomedical and Neuroscience Knowledge Graph

Expanding knowledge bases for LLM reasoning, such as integrating biomedical entities in heterogeneous graphs (Himmelstein et al., 2017), has been investigated. Still, academic resources such as Drug Bank (Wishart et al., 2018) are curated for scientific researchers and not designed for LLM-based natural language reasoning.

In the medical domain, LLM-assisted knowledge graph construction frameworks have recently emerged. CNKG (Sarabadani et al., 2026) designed a mathematical modeling framework for stroke-specific knowledge graph construction via LLM-based knowledge extraction. CNKG (Sarabadani et al., 2026) also proposed a dynamic knowledge graph framework that incorporates LLM reasoning for medical diagnosis and personalized treatment recommendation. GDPKG-LLM (RahseparFard et al., 2025) also extended this direction by integrating genes, disease, pharmacogenomics knowledge graphs for cognitive neuroscience. These approaches illustrate the feasibility of employing LLM with biomedical knowledge graph construction and reasoning pipelines.

On the other hand, in the area of cognitive neuroscience, there are many knowledge graphs with structured resources such as NeuroSynth (Yarkoni et al., 2011), Cognitive Atlas (Poldrack et al., 2011), and Brain-Map. These resources represent an extensive collection on brain activation patterns, cognitive ontologies, and experimental data, but there is no controlled study in the literature investigating how the LLM framework of a given architecture interacts with heterogeneous cognitive neuroscience KGs with structured queries.

## 2.3. QA Benchmarks and Evaluation Gaps

Open-domain multi-hop benchmarks such as Hot-potQA (Yang et al., 2018) and Complex Web Questions (Talmor and Berant, 2018) require compositional reasoning but are built primarily over unstructured documents or single knowledge sources. Graph-specific datasets such as LC-QuAD (Dubey et al., 2019) and MetaQA (Zhang et al., 2018) assess structured query generation but are restricted to single-graph environments. None of these benchmarks jointly evaluate multi-KG reasoning, structured query equivalence, hallucination behavior, and performance differences between grounded and ungrounded inference modes.

### 2.4. Discussion and Positioning.

Existing behavior benchmarks mainly focus on either structured query generation (Tsatsaronis et al., 2015), or knowledge graph construction (Sarabadani et al., 2026; RahseparFard et al., 2025) or application specific tasks such as medical inference and systems biology models. We are not aware of any current benchmark which jointly captures hierarchical subgraph structure generation, structured query equivalence, hallucination behavior, and performance differences between grounded and ungrounded inference modes. In addition, the well-known query benchmarks are restricted to single graph environments. Our significance lies in providing a system-level benchmark of the behavior of LLMs and KG construction in an automated multi-KG setting. We nearly exhaustively evaluate the current state-of-the-art frontier and domain specific LLMs on a variety ranks heterogeneous cognitive neuroscience and biomedical knowledge graphs under both KG-aware and KG-agnostic regimes. By sparsely identifying the LLMs that are best suited for the downstream multi-hop subgraph construction while being minimal in hallucination propensity, we propose a principled and statistically robust method for evaluation and selection of LLMs for cross-domain use cases in the "do not match" validation KGs.

## 3. Material and Methods

To avoid bias, we implemented a multi-annotator framework with annotators blind to the model identities during the evaluation. We report Cohen's Kappa coefficient calculations as a measure of inter-rater reliability. We performed pairwise comparisons between each annotator. The Kappas were calculated to be 0.753 between Annotator 1 and Annotator 2, 0.807 between Annotator 1 and Annotator 3 and 0.784 between Annotator 2 and Annotator 3. This suggests a high level of agreement (Kappa > 0.73) between each pair of annotators; the mean Kappa across all pairwise discourses were 0.781 (SD = 0.027).

Subsequently, to assess overall agreement among all three annotators simultaneously beyond chance, we calculate Fleiss' Kappa for the multi-rater setting. The overall Fleiss' Kappa coefficient was calculated as 0.77, further indicating strong concordance between the three experts over the entire set of annotations. A concise summary of these findings is presented at the beginning of the Results section. Three experts independently coded hallucination and reasoning failures based on a qualitative error scheme. Cohen's Kappa coefficient was evaluated on each possible pair of annotators; the overall mean pairwise Kappa is 0.77 (mean +/- SD: 0.77 +/- 0.10). We evaluate whether contemporary large language models (LLMs) can reliably function as structured interfaces to heterogeneous cognitive neuroscience knowledge graphs (KGs). The hypothesis is that frontier general purpose models would excel in surface level factual retrieval, domain specific models would demonstrate superior stability in complex reasoning tasks when provided with explicit graph g-rounding. Our experimental pipeline consists of five stages: (1) benchmark construction, (2) ground-truth extraction from authoritative KGs,

The hypothesis is that frontier general purpose models would excel in surface level factual retrieval, domain specific models would demonstrate superior stability in complex reasoning tasks. (3) dual-mode model evaluation (KG-aware and KG-agnostic), (4) quantitative and qualitative metric computation, and (5) statistical, error, and human analyses.

## 3.1. Knowledge Graph Environment

We use five KGs covering (NLP-based) concepts, relationships, and ontologies, specifically: Pharmacokinetic Drug Bank, NeuroSynth (term-region associations), Cognitive Atlas (cognitive ontology), Brain Map (neuroimaging metadata), and GDPKG-LLM (LLM-derived neuroscience KG). KGs are represented in RDF and hosted in a SPARQL-enabled triple store (e.g., Blaze-graph/Virtuoso), with property-graph versions for compatible ontologies in Neo4j. A mapping layer links entity types, predicates, and other schema components across graphs.

## 3.2. Knowledge Graph Environment

We employ five complementary KGs: Drug Bank (pharmacology), NeuroSynth (term–region asso-ciations), Cognitive Atlas (cognitive ontology), Brain Map (neuroimaging metadata), and GDPKG-LLM (LLM-derived neuroscience KG). RDF-based KGs are deployed in a SPARQL-enabled triple-store (e.g., Blaze-graph/Virtuoso), while property-graph representations are hosted in Neo4j. A mapping layer documents entity types, predicates, and across graphs.

## 3.3. Benchmark Construction

Let the benchmark query set be defined as

$$Q = \{q_1, q_2, \ldots, q_N\}, \; N = 150. \quad (1)$$

Queries are partitioned into: single-hop (50), multi-hop (70), and high-level reasoning (30).

Each query qi is associated with: a gold structured query gi and canonical answer set Ai obtained by executing gi over the authoritative KG(s). Multi-KG queries use predefined union/intersection opera-tors. All ground-truth outputs undergo double expert validation (Cohen's and Fleis's).

## 3.4. Dual Evaluation Modes

Each model M is evaluated under two regimes.

KG-aware:

$$\hat{A}^{(M)} = execute(\hat{g}^{(M)}), \qquad (2)$$

neuro-symbolic environments. where gˆ(M) is the generated SPARQL/Cypher query.

KG-agnostic:

$$\hat{A}_i^{(M)} = M(q_i), \qquad (3)$$

This separation isolates translation errors from inter-nal reasoning and hallucination behavior. Formal metric definitions

Exact Match:

$$EM^{(M)} = \frac{1}{N}\sum_{i=1}^{N} \mathbb{1} [\hat{A}_i^{(M)} = A_i]. \qquad (4)$$

Precision and Recall (set-valued):

$$Precision_i^{(M)} = \frac{\left|\hat{A}_i^{(M)} \cap A_i\right|}{\left|\hat{A}_i^{(M)}\right|},$$

$$Recall_i^{(M)} = \frac{\left|\hat{A}_i^{(M)} \cap A_i\right|}{|A_i|}. \qquad (5)$$

Macro-F1:

$$\mathrm{F}_1^{(M)} = \frac{1}{N}\sum_{i=1}^{N}\frac{2P_iR_i}{P_i + R_i}. \qquad (6)$$

Parse ability:

$$\mathrm{ParseRate}^{(M)} = \frac{1}{\mathrm{N}}\sum_{\mathrm{i=1}}^{\mathrm{N}} \nVdash \left[\mathrm{Parse}(\widehat{\mathrm{g}_{\imath}}^{(\mathrm{M})})\right]. \qquad (7)$$

Result equivalence:

$$\mathrm{EquivRate}^{(M)} = \frac{1}{\mathrm{N}}\sum_{\mathrm{i=1}}^{\mathrm{N}} \nVdash \left[\mathrm{Execute}(\widehat{\mathrm{g}_{\imath}}^{(\mathrm{M})}\right) = [\mathrm{Execute}((\mathrm{g_i})]. \qquad (8)$$

Hallucination rate:

$$\mathrm{HallucRate}^{(M)} = \frac{1}{\mathrm{N}}\sum_{\mathrm{i=1}}^{\mathrm{N}} \nVdash [\exists \mathrm{c} \notin \mathrm{KG}]. \qquad (9)$$

Latency:

$$\mathrm{Latensy}^{(M)} = \frac{1}{\mathrm{N}}\sum_{\mathrm{i=1}}^{\mathrm{N}} t_i^{(M)}. \qquad (10)$$

Statistical Analysis

We compute 95% confidence intervals using boot-strap resampling (10k samples):

$$\mathrm{CI}_{95\%} = \left[\theta^{*(0.025)}, \theta^{*(0.975)}\right]. \qquad (11)$$

Paired comparisons use the Wilcoxon signed-rank test:

$$H_0\text{: } \mathrm{F}_{1i}^{(M_1)} = \mathrm{F}_{1i}^{(M_2)}. \qquad (12)$$

Binary correctness uses McNemar’s test. Multiple comparisons are corrected via Benjamini–Hochberg:

$$p(k) \leq \frac{k}{m}\alpha. \qquad (13)$$

## 3.5. Evaluation Pipeline

The algorithm of the pipeline is as follows:

Algorithm 1 Dual-Mode LLM Evaluation

```
for each model M in models do
  for each question qi in benchmark do
    Ai ← get_ground_truth(qi)
    if mode = KG-aware then
    prompt ← build_sparql_prompt(qi)
    model_output ← call_model(prompt)
    query ←extract_query(model_output)
      if query = NO_QUERY or invalid then
        predicted_answer ← Ø
        kg_result ← None
    else
        kg_result ← execute(query)
        predicted_answer ←synthesize(kg_result)
      end if
    else
      prompt ← build_direct_prompt(qi)
      model_output ← call_model(prompt)
      predicted_answer ← model_output
      query ← None
      kg_result ← None
    end if
    metrics ← evaluate (predicted_answer, Ai)
    log artifacts
  end for
end for
```

## 3.6. Error and Human Evaluation

Errors are categorized into: (1) translation errors, (2) reasoning failures, (3) hallucinations, (4) schema mis match, (5) ambiguity.

For each model, 30 failure cases are manually annotated.

Human evaluation (n=150 responses) assesses correctness (3-point scale), usefulness, and hallucination presence. Inter-annotator agreement is measured via Cohen's κ.

Reproducibility and Ethics

We release benchmark queries, prompts, evaluation scripts, Docker environment, random seeds, and KG access instructions. All prompts, model outputs, and configurations are logged. Licensing constraints (e.g., Drug Bank) are documented. Model outputs are not in-tended as clinical advice

# 4. Results

Before reporting the primary performance, the reliability of qualitative annotations are revis-ited.

| Model | Parseable (%) | Equivalence (%) | EM Simple (%) | EM Multi-hop (%) | EM Reasoning (%) | Avg F1 |
|---|---|---|---|---|---|---|
| GPT-5.1 | 96 | 91 | 95 | 88 | 85 | 0.91 |
| Gemini 2.0 Pro | 92 | 86 | 90 | 72 | 70 | 0.84 |
| Llama 3 | 87 | 72 | 88 | 55 | 50 | 0.71 |
| Qwen | 85 | 70 | 86 | 54 | 48 | 0.69 |
| BioGPT | 89 | 63 | 80 | 45 | 42 | 0.65 |

Table 1. Performance in KG-aware mode

In Section 3, inter-rater agreement was assessed using both pairwise and multi-rater indices. The pairwise Cohen's Kappa coefficients ranged from 0.753 to 0.807 (mean κ = 0.781), and the overall Fleiss' Kappa was 0.77. These values are more than the conventional 0.60 threshold for substantial agreement (Landis & Koch, 1977). Therefore, they confirm that our error classifications and hallucination annotations are sufficiently consistent to support the subsequent analyses.

Table 1 reports performance in KG-aware mode. GPT-5.1 achieves the highest parse ability (96%) and semantic result-equivalence (91%), indicating strong syntactic query formation and relational correctness. Gemini 2.0 Pro follows with moderate degradation under multi-hop tasks, while Llama 3, Qwen, and BioGPT exhibit substantial drops in semantic equivalence despite relatively high parse ability rates.

The gap between parse ability and equivalence reflects semantic join failures in multi-hop reason-ing. While several models produce syntactically valid SPARQL queries, they fail to preserve correct relational bindings across heterogeneous schemas.

Figure 1 illustrates degradation across query tiers. All models exhibit monotonic performance decline as relational complexity increases, but the magnitude differs substantially. GPT-5.1 maintains relatively stable multi-hop reasoning, whereas BioGPT and smaller open models show sharp compositional brittleness.

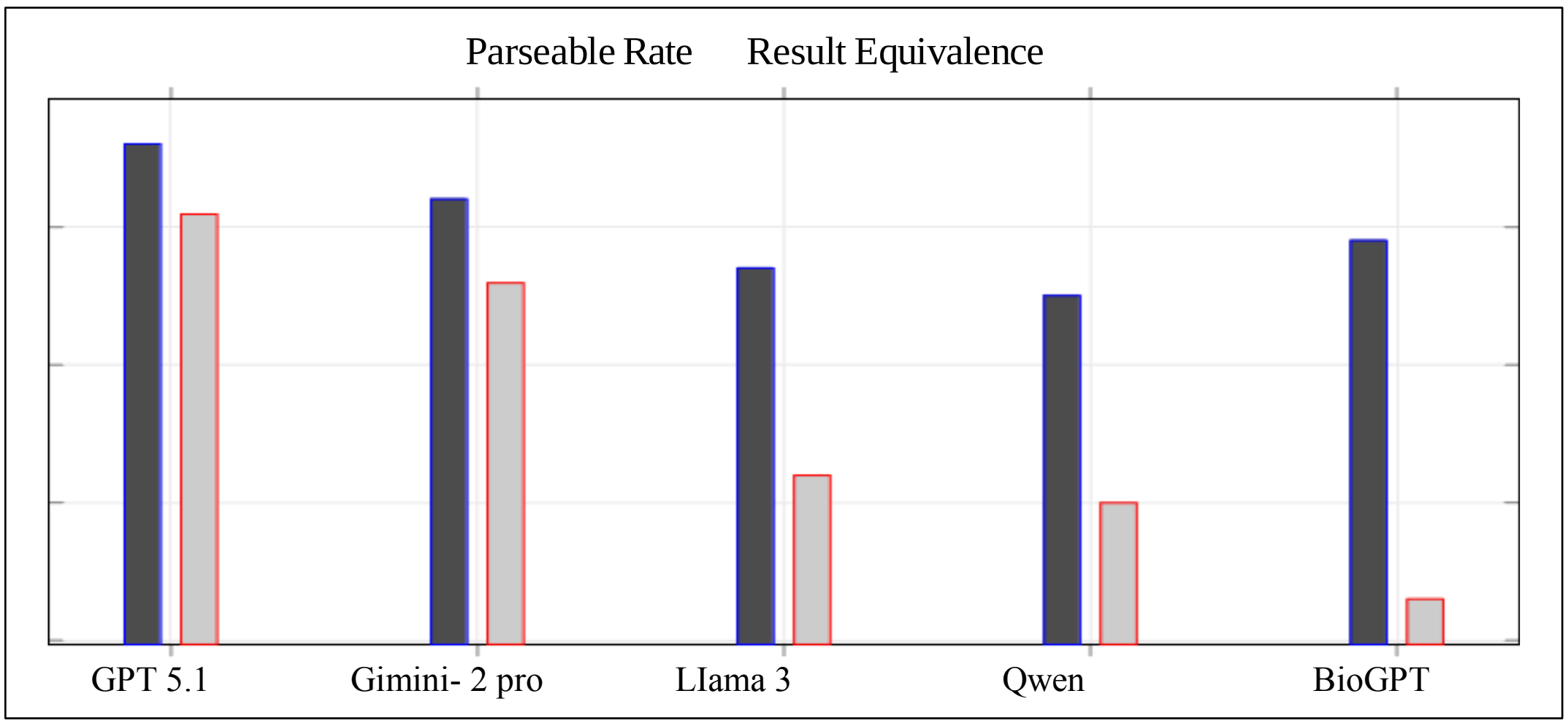

Figure 1: Structural reliability comparison in KG-aware mode. Although several models produce syntactically valid queries, semantic equivalence degrades substantially for smaller models under multi-hop reasoning.

## 4.1. Hallucination and KG-Agnostic Behavior

Table 2 presents performance in KG-agnostic mode. Removing structured grounding leads to systematic hallucination increases. GPT-5.1 maintains the lowest hallucination rate (7%) and highest semantic similarity (0.91), whereas BioGPT reaches 30% hallucination rate, particularly on cross-domain queries.

Table 2. Performance in KG-agnostic mode

| Model | Hallucination (%) | EM Simple (%) | EM Multi-hop (%) | EM Reasoning (%) | Cosine Similarity |
|---|---|---|---|---|---|
| GPT-5.1 | 7 | 92 | 80 | 78 | 0.91 |
| Gemini 2.0 Pro | 15 | 88 | 68 | 65 | 0.84 |
| Llama 3 | 22 | 85 | 55 | 52 | 0.72 |
| Qwen | 18 | 83 | 54 | 50 | 0.70 |
| BioGPT | 30 | 78 | 40 | 38 | 0.68 |

Figure **2** demonstrates a negative correlation be-tween hallucination rate and semantic similarity. Mod-els with higher hallucination rates systematically exhibit lower embedding-level agreement with ground truth answers.

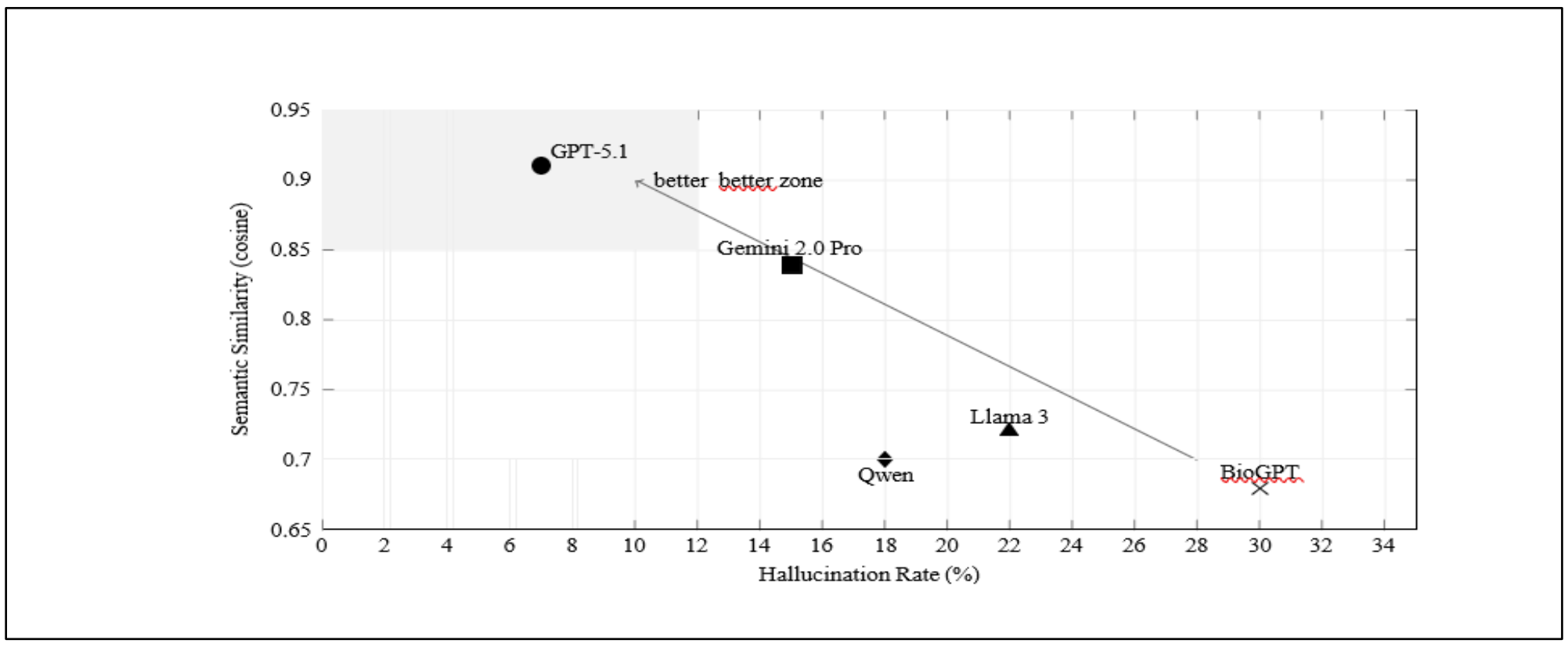

Figure 2: Hallucination frontier in KG-agnostic mode. Models are better when they move toward the upper-left: fewer hallucinations and higher semantic similarity.

## 4.2. Latency–Accuracy Trade-off

Latency results are summarized in Table 3. Fron-tier API-based models incur moderate latency over-head but deliver substantially higher reasoning reliability. Local models are computationally efficient but exhibit reduced compositional robustness.

| Model | KG-aware | KG-agnostic |
|---|---|---|
| GPT-5.1 | 1.5 | 1.2 |
| Gemini 2.0 Pro | 1.8 | 1.4 |
| Llama 3 | 0.4 | 0.3 |
| Qwen | 0.5 | 0.4 |
| BioGPT | 0.45 | 0.35 |

Table 3. Average latency per query (seconds)

Overall, three consistent patterns emerge:

Structured grounding substantially reduces hallucination.

Multi-hop reasoning remains the dominant performance bottleneck.

Syntactic query validity does not guarantee semantic equivalence.

Frontier general-purpose models demonstrate stronger cross-KG relational abstraction, whereas domain-specialized models retain narrow factual strength but limited compositional generalization.

# 5. Discussion

The high inter-rater agreement in the annotation phase lends strong empirical sup-port to our er-ror taxonomy. These levels of consistency satisfactory with standards in the field of (NLP). This annotator concordance is important for our hallucination analysis. It ensures that the differences observed across models genuinely reflect architectural variations in hallucination vulnerability rather than artifacts of subjective annotation bias. The experts in our setup agreed far more than we expected. It suggests its potential utility for standardized evaluation protocols in future LLM-KG benchmarking efforts. Examined how different large language models perform when answering cognitive neuro-science questions using heterogeneous knowledge graphs.

The results show that performance depends not only on model size, but also on training back-ground and access to structured knowledge.

GPT-5.1 consistently achieved the best overall performance. It showed strong query translation, stable multi-hop reasoning, and low hallucination rates. Its performance stayed stable even as questions grew more complex suggesting that large general-purpose models, combined with structured knowledge graphs, can handle complex cross-domain reasoning effectively.

Gemini 2.0 Pro also performed well in KG-aware mode. However, its performance dropped more notice-ably in KG-agnostic mode. Without a knowledge graph, reasoning becomes less accurate and less reliable. Domain-specific models showed a different pattern.

BioGPT performed well on drug-related questions but struggled when tasks required integration across cognitive and neuroimaging knowledge. This indicates that domain-focused training im-proves precision within a narrow area but may limit broader reasoning.

Llama 3 and Qwen support simple factual questions well. But struggled with multi-hop and reasoning tasks. Multi-hop reasoning remains the most challenging setting. As relational depth increased, all models showed some degree of performance decline. GPT-5.1 showed the smallest decline, suggesting stronger in-ternal representation of relational dependencies. Other models often failed to correctly connect entities across multiple knowledge graphs, leading to incomplete or partially incorrect answers. Hallucination analysis further emphasizes the im-portance of knowledge graph grounding.

In KG-agnostic mode, hallucination rates increased for all models. When responses were con-strained by structured graph execution, hallucination decreased significantly. This confirms that structured external knowledge helps stabilize model outputs and reduces unsupported claims. For real-world neuroscience applications, grounding LLMs in validated knowledge graphs is therefore essential.

These findings have practical implications. LLM–KG systems can support complex neuro-scientific queries, cross-dataset integration, and exploratory analysis. However, model choice and ground-ing strategy must be carefully considered. A hybrid approach that combines general-purpose reasoning capacity with domain-specific knowledge may provide a balanced solution. We should note a few limitations. Our benchmark includes five main knowledge graphs.

Performance could change with future updates or fine-tuning. More study should explore reasoning transparency, uncertainty estimation and integration with experimental neuroimaging pipelines.LLMs show real promise for heterogeneous neuroscience knowledge graphs, but to be reliable, they need structured grounding and stronger multi-hop reasoning.

## 6. Conclusion

This study presented a systematic evaluation of five large language models interacting with five heterogeneous cognitive neuroscience and biomedical knowledge graphs. By jointly analyzing query translation, multi-hop reasoning, hallucination behavior, and latency, we provide a detailed view of how current LLMs perform in neuro-symbolic settings. The results show a good pattern. General-purpose models with broad pretraining, particularly GPT-5.1, demonstrate more stable performance across structured and unstructured evaluation modes. They maintain higher translation accuracy, stronger multi-hop reasoning, and lower hallucination rates. In contrast, domain-specific models such as BioGPT perform well within narrow factual do-mains but struggle when reasoning requires integration across multiple graphs or conceptual hierarchies. Good cross-domain reasoning needs more than domain expertise and it needs structural flexibility and general semantic abstraction.

Each model when a query included multi-hop reasoning in our tests broke down. Otherwise, we saw the same pattern for all models. The more complex the relations, it is the worse the performance. This concludes that structured inference over heterogeneous graphs is still a real struggle for today's LLMs.

The experiments further demonstrate that grounding models in validated knowledge graphs significantly reduces hallucination and im-proves reliability. Therefore, structured access to external knowledge should be considered a core component of neuro-symbolic systems rather than an optional enhancement.

Beyond model comparison, this work contributes a reproducible benchmark framework for evaluating LLM–KG interaction in cognitive neuroscience. The dual evaluation setup, combining KG-aware and KG-agnostic modes, helps separate reasoning ability from parametric memorization and offers a clearer under-standing of model behavior.

Future research can extend this work in several directions. More knowledge graphs, fine-tuned neuroscience models and adaptive grounding could deepen our understanding of cross-domain reasoning.

Research on explainability, uncertainty estimation and structured confidence reporting is necessary for real-world application. Integrating LLM–KG systems with experimental neuroimaging or electrophysiological data could enhance practical applicability. Finally, controlled LLM pipelines for incrementally enriching knowledge graphs could help neuroscience generate hypotheses at scale and keep knowledge continuously updated.

Overall, LLMs show real promise for working with heterogeneous neuroscience knowledge graphs, but to be consistently reliable, they need structured grounding and better multi-hop reasoning.